# Similarities and Differences between Machine Learning and Traditional Advanced Statistical Modeling in Healthcare Analytics

Michele Bennett, PhD; Karin Hayes, Ewa J. Kleczyk, PhD; Rajesh Mehta, RPh

## Abstract

Data scientists and statisticians are often at odds when determining the best approach – machine learning or statistical modeling – to solve an analytics challenge. However, machine learning and statistical modeling are more cousins than adversaries on different sides of an analysis battleground.  Choosing between the two approaches or in some cases using both is based on the problem to be solved and outcomes required as well as the data available for use and circumstances of the analysis.  Machine learning and statistical modeling are complementary, based on similar mathematical principles, but simply using different tools in an overall analytics knowledge base.   Determining the predominant approach should be based on the problem to be solved as well as empirical evidence, such as size and completeness of the data, number of variables, assumptions or lack thereof, and expected outcomes such as predictions or causality. Good analysts and data scientists should be well versed in both techniques and their proper application, thereby using the right tool for the right project to achieve the desired results.

## Declaration

- Authors are affiliated with Symphony Health, a division of Icon, plc.
- Dr. Kleczyk is also an Affiliated Graduate Faculty in the School of Economics atThe University of Maine, Orono, Maine
- Dr. Bennett is also an Adjunct Professor, within the Graduate and Doctoral Programs in Data Science, Computer Science, and Business Analytics at Grand Canyon University
- Competing Interest: The authors declare that they have no competing interests.
- Funding: Authors work for Symphony Health, ICON plc Organization.
- Acknowledgements: Authors would like to recognize Heather Valera for her review of document drafts, and valuable feedback in improving the article content

## Introduction

In the recent years, machine learning techniques have been utilized to solve problems at hand across multitudes of industries and topics. In the healthcare industry, these techniques are often applied to a variety of healthcare claims and electronic health records data to garner valuable insights into diagnostic and treatment pathways in order to help optimize patient healthcare access and treatment process (Beam & Kohane, 2018). Unfortunately, many of these applications resulted in inaccurate or irrelevant research results, as proper research protocols were not fully followed (Shelmerdine et al., 2021). On the other hand, statistics has been the basis of analysis in healthcare research for decades, especially, in the areas of clinical trials and health economics and outcomes research (HEOR), where the precision and accuracy of analyses have been the primary objectives (Romano & Gambale, 2013). Furthermore, the classical statistics methodologies are often preferred in those research areas to ensure the ability to replicate and defend the results and ultimately, the ability to publish the research content in peer-reviewed medical journals (Romano & Gambale, 2013). The increased availability of data, including data from wearables, provided the opportunity to apply a variety of analytical techniques and methodologies to identify patterns, often hidden, that could help with optimization of healthcare access as well as diagnostic and treatment process (Razzak et al., 2020).

With the rapid increase in data from the healthcare and many other industries, it is important to consider how to select well- suited statistical and machine learning methodologies that would be best for the problem at hand, the available data type, and the overall research objectives (Bzdok et al., 2018). Machine learning alone or complemented by statistical modeling is becoming, not just a more common, but a desired convergence to take advantage of the best of both approaches for advancing healthcare outcomes (Beam & Kohane, 2018).

## Machine Learning Foundation is in Statistical Learning Theory

Machine learning (ML) is grounded in statistical learning theory (SLT), which provides the constructs used to create prediction functions from data. SLT formalizes a model that makes a prediction based on observations (i.e., data) and ML automates the modeling (von Luxburg & Scholkopf, 2011). The foundation for SLT is multivariate statistics and functional analysis (Bousquet et al., 2003). Functional analysis is the branch of statistics that measures shapes, curves, and surfaces, extending multivariate vector statistics to continuous functions (Bousquet et a., 2003).

In ML, terms such as underfitting and overfitting are often used to describes models that do not generalize the data effectively and might not present the right set of data elements to explain the data patterns and posted hypothesis (Field, 2012). Underfitting is often defined as a model which is missing features that would be present in the most optimized model, akin to a regression model not fully explaining all of the variance of the dependent variable (Field, 2012). In a similar vein, overfitting is when the model contains more features or different features than is optimal, like a regression model with autocorrelation or multicollinearity (Field, 2012). Often dimension reduction approaches like Principal Component Analysis (PCA) or boot strapping techniques as well as subject matter expertise and clinical knowledge can help with resolving some of these concerns and limiting the potential model issues. Furthermore, understanding the studied population and data characteristics can further help define the data to be used, variable selection, and proper model set up (Carmichael, & Marron, 2018).

## Similarities between Machine Learning and Statistical Modeling

There are similarities between ML and statistical modeling that are prevalent across most modeling efforts. For example, analysis starts with the assumption that data or observations from the past can be used to predict the future (Luxburg & Scholkopf, 2011). Variables are generally of two types – dependent variables, that in ML are called targets, and independent variables that in ML are called features. The definition of the variables is the same as in statistical analysis (Bousquet et a., 2003) There is a need for models to be created and the data used in a way that allow for generalization (Luxburg & Scholkopf, 2011) Loss and risk are described frequently in terms of mean squared error (MSE). In statistics and ML, MSE is the difference between the predicted value and the actual value and is used to measure loss of the performance of predictions (Field, 2012).

## Differences between Machine Learning and Statistical Modeling

Differences are distinct and based on purpose and need for the analysis as well as the outcomes. Assumptions and purposes for the analysis and approach can differ. For example, statistics typically assumes that predictors or features are known and additive, models are parametric, and testing of hypotheses and uncertainty are forefront (Breiman, 2001); ML does not make these assumptions. In ML, many models are based on non-parametric approaches where the structure of model is not specified or unknown, additivity is not expected, and assumptions about normal distributions, linearity or residuals, for example, are not needed for modeling (Carmichael & Marron, 2018).

The purpose of ML is predictive performance using general purpose learning algorithms to find patterns that are less known, unrelated, and in complex data without *a priori* view of underlying structures (Carmichael & Marron, 2018). Whereas in statistical modeling,

5consideration for inferences, correlations, and the effects of a small number of variables are drivers (Breiman, 2001).

Machine learning is highly effective when the model uses more than a handful of independent variables/features (Carmichael & Marron, 2018). ML is required when the number of features (p) is larger than the number of records or observations (n) – this is called the *curse of dimensionality* (Bellman, 1961), which does increase the risk of overfitting, but can be overcome with dimensionality reductive techniques (i.e., PCA) as part of modeling (Bzdok et al., 2017) and clinical/expert input on the importance or lack thereof of certain features is it relates to the disease or its treatment. Additionally, statistical learning theory teaches that learning algorithms increase their ability to translate complex structures from data at a greater and faster rate than the increase of sample size capture can alone provide (Bousquet et al., 2003). Therefore, statistical learning theory and ML offer methods for addressing high-dimensional data or big data (high velocity, volume and variety) and smaller sample sizes (Hastie et al., 2016) such as recursive feature elimination and support vector machines (Qiu et al., 2014), boosting, or cross validation which can also minimize prediction error (Chapman et al., 2016).

Model validation is an inherent part of the ML process where the data is split into training data and test data, with the larger portion of data used to train the model to learn outputs based on known inputs. This process allows for rapid structure knowledge for primary focus on building the ability to predict future outcomes (Bzdok et al., 2017). Beyond initial validation of the model within the test data set, the model should be further tested in the real world using a large, representative, and more recent sample of data (Argent et al., 2021). This can be accomplished by using the model to score the eligible population and using a look forward period to assess incidence or prevalence of the desired outcome. If the model is performing well, probability scores should be directly correlated to incidence/prevalence (the higher the probability score, the higher the incidence/prevalence). Model accuracy, precision, and recall can also be assessed using this approach (Parikh et al., 2008).

## How to Choose between Machine Learning and Statistical Modeling

Machine learning algorithms are a preferred choice of technique vs. a statistical modeling approach under specific circumstances, data configurations, and outcomes needed (See Figure 2).

*Importance of Prediction over Causal Relationships*

Machine learning algorithms are leveraged for prediction of the outcome rather than present the inferential and causal relationship between the outcome and independent variables / data elements (Hastie et al., 2016; Hayes et al, 2019). Once a model has been created, statistical analysis can sometime elucidate the importance and relationship between independent and dependent variables.

*Application of Wide and Big Dataset(s)*

Machine Learning algorithms are learner algorithms and learn on large amount of data often presented by a large number of data elements, but not necessarily with many observations (Belabbas & Wolfe, 2009). Ability of multiple replications of samples, cross validation or application of boot strapping techniques for machine learning allows for wide datasets with many data elements and few observations, which is extremely helpful in predicting rare disease onset (Kempa-Liehr et al., 2020) as long as the process is accompanied with real world testing to ensure the models are not suffering from overfitting (Argent et al. 2020; Chapman et al., 2016). With the advent of less expensive and more powerful computing power and storage, multi-algorithm, ensembled models using larger cohorts can be more efficiently built. Larger modeling samples that are more representative of the overall population can help reduce the likelihood of overfitting or underfitting (Wasserman, 2013).

Statistical models tend to not operate well on very large datasets and often require manageable datasets with a fewer number of pre-defined attributes / data elements for analysis (Belabbas & Wolfe, 2009). The recommended number of attributes is up to 12 in a statistical model, because these techniques are highly prone to overfitting (Wasserman, 2013).

*Limited Data and Model Assumptions Are Required*

In machine learning algorithms, there are fewer assumptions that need to be made on the dataset and the data elements (Bzdok et al. 2018). However, a good model is usually preceded by profiling of the target and control groups and some knowledge of the domain. Understanding relationships within the data improve outcomes and interpretability (Child & Washburn, 2019).

Machine learning algorithms are comparatively more flexible than statistical models, as they do not require making assumptions regarding collinearity, normal distribution of residuals, etc. (Bzdok et al., 2018). Thus, they have a high tolerance for uncertainty in variable performance (e.g., confidence intervals, hypothesis tests (Hillermeir & Waegerman, 2021). In statistical modeling emphasis is put in uncertainty estimates, furthermore, a variety of assumptions have to be satisfied before the outcome from a statistical model can be trusted and applied (Hullermeier & Waegerman, 202l). As a result, the statistical models have a low uncertainty tolerance (Wasserman, 2013).

Machine learning algorithms tend to be preferred over statistical modeling when the outcome to be predicted does not have a strong component of randomness, e.g., in visual pattern recognition an object must be an E or not an E (Bzdok et al., 2018), and when the learning algorithm can be trained on an unlimited number of exact replications (Goh et a., 2020)

ML is also appropriate when the overall prediction is the goal, with less visibility to describe the impact of any one independent variable or the relationships between variables (Chicco, & Jutman, 2020), and when estimating uncertainty in forecasts or in effects of selected predictors is not a requirement (Hillermeir & Waegerman, 2021). However, often data scientists and data analysts leverage regression analytics to understand the estimated impact, including directionality of the relationships between the outcome and data elements, to help with model interpretation, relevance, and validity for the studied area (Child & Washburn, 2019).

ML is also preferred when the dataset is wide and very large (Belabbas & Wolfe, 2009) with underlying variables are not fully known and previously described (Bzdok, et al., 2018).

## Machine Learning Extends Statistics

Machine learning requires no prior assumptions about the underlying relationships between the data elements. It is generally applied to high dimensional data sets and does not require many observations to create a working model (Bzdok, et al., 2018). However, understanding the underlying data will support building representative modeling cohorts, deriving features relevant for the disease state and population of interest, as well as understanding how to interpret modeling results (Argent et al., 2021; Child & Washburn, 2019).

In contrast, statistical model requires a deeper understanding how the data was collected, statistical properties of the estimator (p-value, unbiased estimators), the underlying distribution of the population, etc. (Haste, et al., 2016). Statistical modeling techniques are usually applied to low dimensional data sets (Wasserman, 2013).

## Machine learning can Extend the Utility of Statistical Modeling

Robert Tibshirani, a statistician and machine learning expert at Stanford University, calls machine learning "glorified statistics," which presents the dependence of machine learning

9techniques on statistics in a successful execution that not only allows for a high level of prediction, but interpretation of the results to ensure validity and applicability of the results in the healthcare industry (Hastie et al., 2016). Understanding the association and knowing their differences enables data scientists and statisticians to expand their knowledge and apply variety of methods outside their domain of expertise. This is the notion of "data science," which aims to bridge the gap between the areas as well as bring other important to consider aspects of research (Bzdok et al., 2018). Data science is evolving beyond statistics or more simple ML approaches to incorporate self-learning and autonomy with the ability to interpret context, assess and fill in data gaps, and make modeling adjustment over time (Ansari et al., 2018). While these modeling approaches are not perfect and more difficult to interpret, they provide exciting new options for difficult to solve problems, especially where the underlying data or environment is rapidly changing (Child & Washburn, 2019).

Collaboration and communication between not only data scientists and statisticians but also medical and clinical experts, public policy creators, epidemiologists, etc. allows for designing successful research studies that not only provide predictions and insights on relationships between the vast amount of data elements and health outcomes (Chicco & Jutman, 2020), but also allow for valid, interpretable and relevant results that can be applied with confidence to the project objectives and future deployment in the real world (Chicco & Jutman, 2020; Morganstein et al., 2020).

Finally, it is important to remember that machine learning foundations are based in statistical theory and learning. It may seem machine learning can be done without a sound statistical background, but this leads to not really understanding the different nuances in the data and presented results (Hastie et al., 2016). Well written machine learning code does not negate the need for an in-depth understanding of the problem, assumptions, and the importance of interpretation and validation (Goh et al., 2020).

## Specific Examples in Healthcare

As mentioned earlier in the article, machine learning algorithms can be leveraged in the healthcare industry to help evaluate a continuum of access, diagnostic and treatment outcomes, including prediction of patient diagnoses, treatment, adverse events, side effects, and improved quality of life as well as lower mortality rates (Kempa-Liehr et al., 2020).

Often these algorithms can be helpful in predicting a variety of disease conditions and shortening the time from awareness to diagnosis and treatment, especially in rare and underdiagnosed conditions, estimate the 'true' market size, predicting disease progression such as identifying fast vs. slow progressing patients as well as determinants of suitable next line change (Morganstein et al., 2020). Finally, the models can be leveraged for patient and physician segmentation and clustering to identify appropriate targets for in-person and non-personal promotion. (Chicco & Jutman, 2020).

There are, however, instances in which machine learning might not be the right tool to leverage, including when the condition or the underlying condition have a few known variables, when the market is mature and has known predetermined diagnostic and treatment algorithm, and when understanding correlations and inference is more important than making prediction (Bzdok et al., 2018).

**Figure 1:** *Examples of Machine Learning Applications in Healthcare Analytics* (Hayes, Rajabathar, & Balasubramaniam, 2019).

| Predicting Disease Onset | Estimating "True" Market Size | Predicting Disease Progression | Patient/Physician Segmentation & Targeting | Think Twice Before Using Machine Learning |
|---|---|---|---|---|
| ▸ Rare conditions with genetic predispositions<br>▸ Highly heterogenous disease<br>▸ Asymptomatic/ indolent diseases<br>▸ Underdiagnosed conditions<br>▸ Misdiagnosed conditions | ▸ Underdiagnosed conditions<br>▸ Ambiguous diagnosis<br>▸ Treatments with prior event qualifications<br>▸ Analog markets | ▸ Fast & slow progressing conditions<br>▸ Determinations of most suitable next line of treatment<br>▸ Treatment recommendations<br>▸ Abnormal imaging studies/findings or lab tests | ▸ Most suitable patient profiles for therapy<br>▸ Similar physician treating profiles | ▸ Few known variables<br>▸ Correlations are more important than predictions<br>▸ Forecasting in a mature market with no or few changes in market dynamics<br>▸ Conditions with determined diagnosis pathway or clear test results |

One aspect of the machine learning process is to involve a cross functional team of experts in the healthcare area to ensure that the questions and problem statement along with hypothesis are properly set up (Terranova et al., 2021). Many therapeutic areas require in-depth understanding of the clinical and medical concepts (i.e., diagnostic process, treatment regimens, potential adverse effects, etc.), which can help with the research design and selection of the proper analytical techniques. If the expert knowledge is not considered or properly captured in the research design, it might lead to irrelevant, invalid, and biased results, and ultimately invalidate the entire research study (Terranova et al., 2021).

### A Practical Guide to the Predominant Approach

Using a real example of a project with the goal of predicting the risk of hypertension due to underlying comorbid conditions or induced by medication, the decision to lead with machine learning vs. statistical modeling can be based on explicit criteria that can be weighed and ranked based on the desired outcome of the work (Hastie et al., 2016; Morganstein et al., 2020).

**Figure 2:** *Criteria for Choosing the Predominant Approach for a Project*

| | Criteria | Project | Predominant Approach (Statistical — Combined — ML) |
|---|---|---|---|
| 1 | Problem to be solved | Predicting hypertension is complicated by numerous related and unrelated health problems, treatment and patient characteristics. High dimensional feature space is amenable to machine learning approaches | ML |
| 4 | Market shifts and complexity | Variability and severity of symptoms, comorbidities, multiple new and existing treatment regimens, and side effects of narcolepsy requires statistical approaches to explore cohorts, and design custom variables | Statistical |
| 2 | Big data and required computing power | 1000s of variables across multiple large claims data sets required for modeling, real world validation, and population scoring. ML models can be run and iterated in an efficient and timely manner | ML |
| 3 | Predictive Power-vs-Explanation of correlations | While highly predictive, ML models alone don't explain relationships between independent variables and the predicted outcome. Statistical approaches describe correlations and are used to tune model features | Combined |
| 5 | Exploratory vs Confirmatory | New and unexpected important features/variables can be discovered through ML modeling. Statistical analyses + clinical expertise used to validate and further describe findings | Combined |

## Conclusion

Machine learning requires fewer assumptions about the underlying relationships between the data elements. It is generally applied to high dimensional data sets and require fewer observations to create a working model (Bzdok, et al., 2018). In contrast, statistical model requires an understanding of how the data was collected, statistical properties of the estimator (p-value, unbiased estimators), the underlying distribution of the population, etc. (Haste, et al., 2016). Statistical modeling techniques are usually applied to low dimensional data sets (Wasserman, 2013). Statistical modeling and ML are not at odds but rather complementary approaches that offer choice of techniques based on need and desired outcomes. Data scientists and analysts should not necessarily have to choose between either machine learning or statistical modeling as a mutually exclusive decision tree. Instead, selected approaches from both areas should be considered as both types of methodologies are based on the same mathematical principles but expressed somewhat differently (Bzdok, et al., 2018; Carmichael & Marron, 2018).